\documentclass{llncs}
\usepackage{amsmath,amssymb}
\usepackage{graphicx}

\begin{document}
\title{Evolution and Computational Learning Theory: A survey on Valiant's paper}

\subtitle{}

\author{Arka Bhattacharya}

\institute{Columbia University,\ New York\ \email{\it (ab3899@columbia.edu)}}

\maketitle

\begin{abstract}
Darwin's theory of evolution is considered to be one of the greatest scientific gems in modern science. It not only gives us a description of how living things evolve, but also shows how a population evolves through time and also, why only the fittest individuals continue the generation forward. The paper basically gives a high level analysis of the works of Valiant[1]. Though, we know the mechanisms of evolution, but it seems that there does not exist any strong quantitative and mathematical theory of the evolution of certain mechanisms. What is defined exactly as the fitness of an individual, why is that only certain individuals in a population tend to mutate, how computation is done in finite time when we have exponentially many examples: there seems to be a lot of questions which need to be answered. [1] basically treats Darwinian theory as a form of computational learning theory, which calculates the net fitness of the hypotheses and thus distinguishes functions and their classes which could be evolvable using polynomial amount of resources. Evolution is considered as a function of the environment and the previous evolutionary stages that chooses the best hypothesis using learning techniques that makes mutation possible and hence, gives a quantitative idea that why only the fittest individuals tend to survive and have the power to mutate.
\end{abstract}

\section{Introduction and Basic Definitions}
Darwin said, evolution consists of many complex mechanisms, which can come into existence without any unlikely events to occur. Evolution consists of a path consisting of many interdependent small stages, but what are the conditions which make these paths to be taken and others, not? It may take exponential time, if evolution just randomly searched all the possible paths, but since we know that the time to evolve is a large finite polynomially bounded value it means there may exist an efficient learning mechanism, which can learn certain function classes and cannot learn others. So, basically mechanisms are treated as mathematical functions in this analysis, such that some functions can be learned in polynomial time, but others cannot due to their inherent computational intractability. We describe some notions, which may help us to formally analyze the quantitative theory of evolvability. We all know, that a cell consists of various types of proteins and maybe, other chemicals and complex circuits and thus, its working depends on many variables. So, in order to define some formalized theory of evolution, mechanisms need to be represented as many argument functions. Hence, a mechanism can take a lot of input parameters to properly function. So, what are many argument functions and how do we represent a mechanism through many argument functions?  Well, a many argument function is a function $f$, which takes more than one input parameters to produce an output. The complex structures of living cells, have to respond to wide variations in both external and internal conditions. Say the conditions are represented by $n$ boolean variables, $x_{1},x_{2},...,x_{n}$ and let us have a function, say $f$ whose output shows some particular desirable response under a particular combination of the $x_{i}^{'}$s, for all $i\in [n]$. We say, the function $f(x_{1},x_{2},...,x_{n})$ is an ideal function and since, it depends on many input parameters, so it is a many argument function.Later in the text, it is explicitly shown that the class of parity functions is not evolvable, while the class of monotone conjunctions over the uniform distribution is. Let $X_{n}$ be the set of all $2^{n}$ possible values that all the $x_{i}^{'}$s can take. Let $D_{n}$ be the probability distribution over $X_{n}$, which basically gives the relative frequency of the occurrence of certain combinations of the $x_{i}^{'}$s.Now, let us have the notion of performance.\newline \newline
\textit{Definition 1.1.}\ Let us have a function $r:X_{n}\rightarrow \{-1,1\}$. The performance of function $r$ with respect to the ideal function $f:X_{n}\rightarrow \{-1,1\}$ for the probability distribution $D_{n}$ over $X_{n}$ is 
\begin{equation}
P_{f}(r,D_{n})=\sum_{x\in X_{n}}f(x)r(x)D_{n}(x)
\end{equation}
Say for some points $x\in X_{n}$ which have non-zero probability in $D_{n}$, we have $\sum_{x\in X_{n}:D_{n}(x)\not=0}D_{n}(x)=1$. Now, if the ideal function $f$ completely agrees with $r$ on these set of points, then we have $f(x).r(x)=1$ for all of them and thus,$P_{f}(r,D_{n})=\sum D_{n}(x)\ =1$. Again, if $f$ does not agree with $r$ for all these $x$, then we have $f(x).r(x)=-1$ and so $P_{f}(r,D_{n})=-\sum D_{n}(x)\ =-1$.So, the range of $P_{f}(r,D_{n})$ is $[-1,1]$ and can be viewed as a fitness landscape over all the genomes $r$.All the points in $X_{n}$ can be thought of as life experiences. When $r$ agrees with the ideal function, we have a benefit, otherwise we have a penalty in case of disagreement.So, over a sequence of life experiences, the organisms or groups which have high values of the performance function are selected preferentially for survival over organisms which have low values of the performance function. It is thus, a basic mathematical definition of the Darwinian concept of \textit{Survival of the Fittest}. An organism or a group can test the performance of its genome $r$ against the ideal function, by sampling a set $S\subset X_{n}$, of poly(n) size, say $s(n)$ life experiences. Let us have a definition concerning empirical performance, which concerns the size of the independent selections, $s$.\newline \newline
\textit{Definition 1.2.}\ The empirical performance, $P_{f}(r,D_{n},s)$, for some positive integer $s$ is a random variable which makes $s$ independent selections from the set $S$ with replacements according to the distribution $D_{n}$ and has the value $\frac{1}{s}.\sum_{x\in S}f(x)r(x)$. \newline \newline
We take $s(n)$ to be the upper bound of population size, as the life experiences $x_{i}$ may correspond to one or more organisms.Moreover, let us also insist that evolution is able to proceed from any starting point, otherwise proceeding back to the reinitialized state from the current state may heavily decrease the value of the performance function.Let us discuss in short, the final two notions before proceeding to the definition of the concept of evolvability.Since, the organisms that can exist at any time is finite and is polynomially bounded, so for a function only a limited number of variants can be explored per generation, whether through mutations or recombination.And finally, we say that mechanisms with significant improvements in the value of performance function, evolve in a limited number of generations.\newline \newline
Let us now have an idea of evolvability and some of its definitions, in terms of learning theory.
\section{Evolvability and its definitions}
From the perspective of learning theory, we can ask whether there exists a hypothesis $r$ for a target concept $f$, such that $r$ closely approximates $f$, the ideal function.Let the concept class $C$ consists of all the ideal functions. Since, evolution has a path and follows a finite number of steps, before it tries to be as close to $f$ as possible, so let us say that the hypothesis for the initial stage be $r_{1}$, the hypothesis for the second stage be $r_{2}$ and continuing like this, the hypothesis for the $i^{th}$ stage is $r_{i}$. The path of evolutionary sequence can be represented by $r_{1}\Rightarrow r_{2}\Rightarrow ...$, where we have $P_{f}(r_{i},D_{n})>P_{f}(r_{i-1},D_{n})$ and each hypothesis tries to approximate the ideal function better than its predecessor. If after a certain number of steps, say $k$, such that the $k^{th}$ hypothesis is $r_{k}$ and we have $P_{f}(r_{k},D_{n})-P_{f}(r_{1},D_{n})\geq d$, for some positive threshold $d$, then the process of evolution starts from any function. Hence, in short we have to get a good evolved representation $r$, from the representation class, say $R_{n}$, such that $P_{f}(r,D_{n})$ is very close to 1.We assume that any representation from $R_{n}$ is polynomially evaluatable, i.e $r(x)$ can be computed in polynomial time, where $x\in X_{n}$.Let the error parameter of the evolved representation be $\epsilon$.\newline \newline
\textit{Definition 2.1.}\ A $p$-neighborhood N on R, for polynomial $p$ and representation class $R$ is a pair of two randomized Turing machines $M_{1}$ and $M_{2}$, such that $M_{1}$ outputs all the $p(n,1/\epsilon)$neighbors of a representation $r\in R_{n}$ so that they all have error parameters of $\epsilon$, using $n$ and $1/\epsilon$ as its input and $M_{2}$ takes all the outputs of $M_{1}$ as its input and returns some representation with probability at least equal to $1/p(n,1/\epsilon)$. \newline \newline
Let the representations, which are generated by $M_{1}$ be put in the set $Neigh_{N}(r,\epsilon)$, so that we have $|Neigh_{N}(r,\epsilon)|=p(n,1/\epsilon)$.\ $M_{1}$ may also do random coin tosses to output members from the set $Neigh_{N}(r,\epsilon)$. After this, $M_{2}$ takes all the members of $Neigh_{N}(r,\epsilon)$ as its input and returns some $r_{c}\in Neigh_{N}(r,\epsilon)$ with probability, $Pr_{N}(r,r_{c})\geq 1/p(n,1/\epsilon)$.Since, the size of $Neigh_{N}(r,\epsilon)$ is polynomially bounded, it means that the number of variants which can be searched is not unlimited, which is logical as the population at any time is finite.\ $M_{2}$ makes sure that differences in performance can be detected reliably.\newline
Most of the exponentially many variants are considered to be impractical and hence, are discarded, while only $poly(n)$ variants are assumed to be feasible for evolution.It can be seen from the analysis, that all the resources, computation time, population and generation sizes are upper bounded by some polynomial, which depends on the number of variables and the inverse of the error parameter and this makes PAC learning possible,which we will analyze later. \newline \newline
To give a basic intuitive understanding of Definition 2.1, consider the diagram below. 
\begin{figure}[ht!]
\centering
\includegraphics[width=90mm]{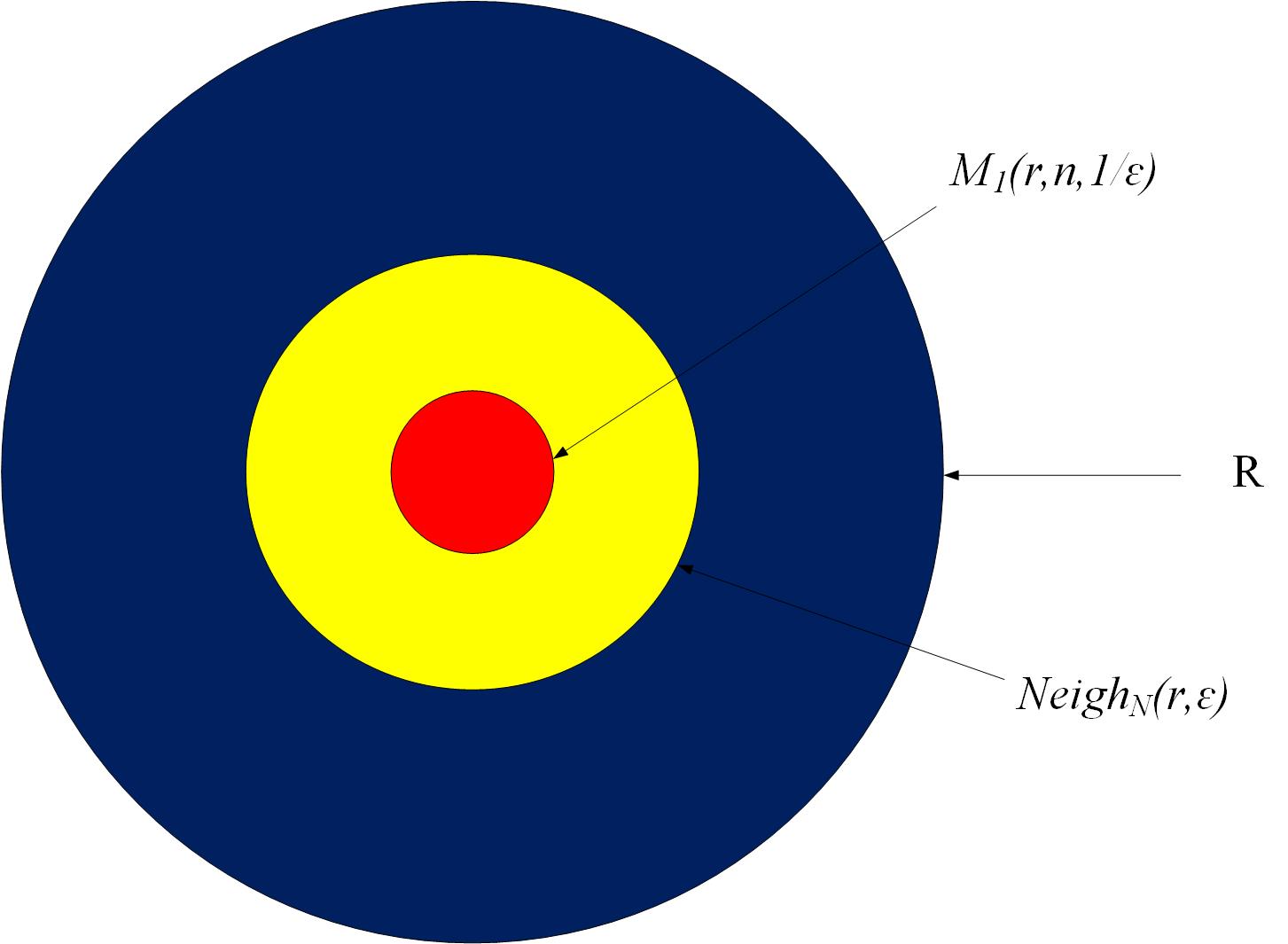}

\end{figure}
The blue region corresponds to the representation class, $R$, the yellow region consists of all the neighboring representations of $r$ having error parameter of $\epsilon$, generated by the randomized Turing Machine, $M_{1}$ and the red region consists of $r$, which is taken as input by $M_{1}$. The yellow region has a size of $p(n,1/\epsilon)$ and $M_{2}$ randomly selects some representation $r_{c}$ from this region with probability at least equal to the inverse of the size of this region. Now, let us define the Mutator function, which helps in genetic mutation. \newline \newline
\textit{Definition 2.2.}\ The mutator function, $M(f,p,R,N,D,s,r,t)$(where all of the input parameters represent same concepts as before and there exists a $p(n,1/\epsilon)$ neighborhood $N$ on $R$) for the current representation, $r\in R_{n}$ is a random variable, which outputs some variant $r_{c}\in Neigh_{N}(r,\epsilon)$ with a certain probability, such that the performance value of $r_{c}$ with respect to the ideal function, $f$ never goes below the performance value of $r$ with respect to $f$ by some threshold $t>0$.\ For some $r_{1},r_{2}\in R_{n}$, if we have $r_{1}\Rightarrow r_{2}$ in the evolutionary path, we say that $r_{2}=M(f,p,R,N,D,s,r_{1},t)$.\newline \newline
The above definition captures the notion of feasibility of an evolutionary path, somewhat in a lower level. Say, for some representation, $r$ the mutator function $M$ computes some variant $r_{c}$ of it and computes the value of the performance, $P_{f}(r_{c},D_{n})=P(r_{c})$.(We will represent the performance of a representation $r$ with respect to the ideal function over a distribution $D$, by $P(r)$ from now onwards). It is easy to see that if $P(r_{c})-P(r)\geq t$, then it is beneficial for evolution and we create a set $POS$, which contains all the positive variants of $r$, capable of mutation. Similarly, let us create a set $NEUT$, to keep all the variants, which have performance values, at least equal to $P(r)-t$, but not exceeding $P(r)$. Hence, $NEUT=\{r_{c}|P(r_{c})\geq P(r)-t\}-POS$.Now, if the cardinality of the set $POS$ is at least equal to 1, we output a representation, $r_{c}\in POS$ with probability equal to $Pr_{N}(r,r_{c})/\sum_{r_{c}\in POS}Pr_{N}(r,r_{c})$, otherwise we output a representation $r_{c}\in NEUT$, with probability $Pr_{N}(r,r_{c})/\sum_{r_{c}\in NEUT}Pr_{N}(r,r_{c})$.We refer to $t$ as the tolerance variable of a representation.Let $t_{i}$ denote the tolerance variable of each representation, $r_{i}$, such that it is generated by a Turing machine on inputs $r_{i-1},n$ and the error parameter $\epsilon$. It is logical to assume that each $t_{i}$ diminishes with $\epsilon$, as we want an accuracy of $1-\epsilon$, whereas population size or the no. of generations is inversely proportional to $\epsilon$. So, we can bound each $t_{i}$ by two polynomially related polynomials $p_{l}$ and $p_{u}$, such that we have $p_{l}(1/n,\epsilon)\leq t_{i}\leq p_{u}(1/n,\epsilon)$.We also have the number of experiences represented by the polynomial $s(n,1/\epsilon)$.So, basically, if we need higher accuracy, then $\epsilon$ needs to be small and tolerance levels,$t_{i}$ should be less, so each variant must be closer to the parent genome with respect to performance.However, we need more life experiences, $s$ for analyzing the conditions of evolvability, which is logical.(at least $s$ is polynomially bounded). So, we have the following equation, representing evolvability of variants.
\begin{equation}
r_{i}=M(f,p(n,1/\epsilon),R_{n},N,D_{n},s(n,1/\epsilon),r_{i-1},t_{i})
\end{equation}
\textit{Definition 2.3.}\ A class $C$ is \textit{tolerance evolvable} iff there exists a polynomial $g(n,1/\epsilon)$(generation size for evolution)and a Turing machine $T$, which computes $t_{i}$ for every $r_{i}\in R_{n}$, where $p_{l}(1/n,\epsilon)\leq t_{i}\leq p_{u}(1/n,\epsilon)$, such that for every positive integer $n$, every $f\in C_{n}$, every $0\leq \epsilon <1$ and every initial representation, $r_{1}\in R_{n}$, the evolution sequence $r_{1},r_{2},r_{3},...$, where $r_{i}=M(f,p(n,1/\epsilon),R_{n},N,D_{n},s(n,1/\epsilon),r_{i-1},t_{i})$ must have a representation, $r_{g(n,1/\epsilon)}$, such that its performance value with respect to $f$ is at least $1-\epsilon$.\newline \newline
\textit{Definition 2.4.}\ A class $C$ is \textit{poly evolvable} over $D$ iff there exists polynomially related variables, $p_{l}$ and $p_{u}$, such that $C$ is \textit{tolerance evolvable} by $p(n,1/\epsilon),s(n,1/\epsilon),R$ and $N$ over $D$.\newline \newline
\textit{Definition 2.5.}\ A class $C$ is \textit{R-evolvable} iff $C$ is \textit{poly evolvable} over $D$ for some polynomials $s(n,1/\epsilon), p(n,1/\epsilon)$ and some $p(n,1/\epsilon)$ neighborhood $N$ on $R$. \newline \newline
\textit{Definition 2.6.}\ A class $C$ is \textit{D-evolvable}, if for some representation class, $R$ it is \textit{R-evolvable} over $D$. \newline \newline
\textit{Definition 2.7.}\ A class $C$ is \textit{perfectly-evolvable} if it is \textit{D-evolvable} over all $D$. \newline \newline
Thus, to give a low level intuitive overview of the definitions, consider the diagram below.(Note that the diagram looks like a decision list, but it is not a decision list by any means. The diagram is just for basic intuitive understanding of the definitions above) From the definitions it is clear that one notion follows from the previous, only if the previous is true. Thus, a concept class which is \textit{perfectly-evolvable} is the broadest of all the notions and is independent of any distribution.\newline
\begin{figure}[ht!]
\centering
\includegraphics[width=135mm]{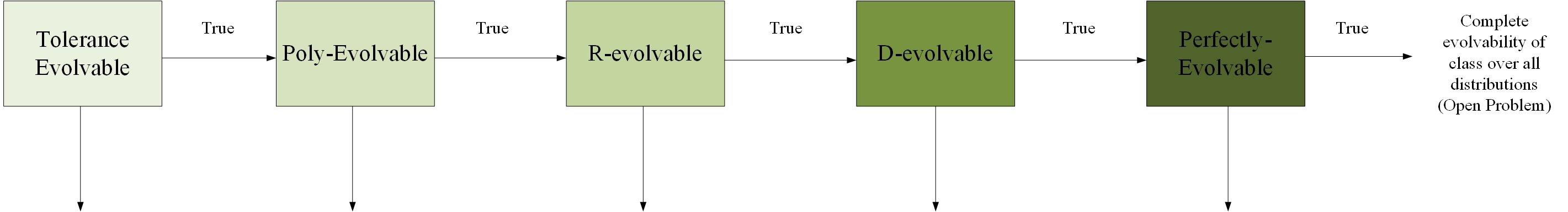}

\end{figure}

\section{Propositions and proofs}
\textit{Proposition 3.1.}\ If there exists a concept class, $C$ which is \textit{R-evolvable} over a distribution $D$, then $C$ is efficiently PAC-learnable by $R$ over $D$.\newline \newline
\textit{Proof.}\ \ 
Recall the basic definition of Probably Approximately Correct (PAC) learning model from [2] and [4].\newline \newline
\textit{Definition.}\ \ If $C$ is a concept class over some domain $X$, then $C$ is PAC learnable if there exists an algorithm $A$, such that for every target concept $c\in C$, for every distribution $D$ on $X$ and for all $0<\epsilon,\delta<1/2$, it outputs a hypothesis $h$ from the hypothesis class $H$ with probability at least $1-\delta$ so that the error of $h$ is upper bounded by $\epsilon$, provided the algorithm is given access to an example oracle $EX(c,D)$ and supplied with the parameters $\epsilon$ and $\delta$.\newline \newline
The above proposition says that the concept class $C$ is \textit{R-evolvable} over $D$, so this means that $C$ is \textit{poly-evolvable} over $D$ for polynomials $s(n,1/\epsilon),p(n,1/\epsilon)$ and some $p(n,1/\epsilon)$ neighborhood $N$ on $R$ and since, it is \textit{poly-evolvable}, this means there exists two polynomially bounded polynomials $p_{l}$ and $p_{u}$, such that $C$ is \textit{tolerance evolvable} by $s,p,R$ and $N$ over $D$. This means there exists some evolutionary sequence $r_{1}\Rightarrow r_{2}\Rightarrow r_{3}\Rightarrow...$, which satisfies equation $(2)$, such that after $k$ evolutionary steps, where $k=O(g(n,1/\epsilon))$, the final hypothesis $r_{k}$ satisfies $P(r_{k})>1-\epsilon$, provided the tolerance values for each representation, $t_{i}$ are generated by some Turing machine $T$, on inputs $r_{i-1},n$ and $\epsilon$.\ The evolution algorithm runs in many stages and is fed with labelled examples from the distribution, $D$. Say, for some stage $i$, let us have a set $S_{i}$ of labelled examples, such that its cardinality, $k_{i}$ is bounded by the polynomial $s(n,1/\epsilon)$. In this case, our target concept is the ideal function, $f$. We have $S_{i}=\{\langle x_{1},f(x_{1})\rangle,\langle x_{2},f(x_{2})\rangle,...,\langle x_{k_{i}},f(x_{k_{i}})\rangle\}$, where $k_{i}=O(s(n,1/\epsilon))$ and the algorithm returns a hypothesis, $r_{i}$ in each stage, $i$ under polynomial time as per our assumptions.Now, the empirical performance of all the possible hypotheses for the current stage is computed in polynomial time and the best one is made the new genome, so that it has the capacity to evolve. Thus, the new hypothesis is generated in polynomial amount of time, using polynomially bounded resources and polynomially bounded input parameters.But, the computation of the performance value for the current hypothesis, is nothing but measuring the probability that the hypothesis matches the target concept on some example point from the distribution $D$. So, for a stage $i$, input domain $X_{n}$ and distribution $D_{n}$, we have $P(r_{i})=\sum_{x\in X_{n}}f(x)r_{i}(x)D_{n}(x)=Pr_{x\sim D}[r_{i}(x)=f(x)]$. Moreover, since the final hypothesis, $r_{k}$ has a performance of at least $1-\epsilon$, where $k=O(g(n,1/\epsilon))$, this means we have $Pr_{x\sim D}[r_{k}(x)=f(x)]\geq 1-\epsilon$, which tells us that the error of the final evolved genome is at most $\epsilon$, i.e $error(r_{k})\leq \epsilon$. All of these statements completely agree with the basic definition of the PAC learning model, provided above and thus, it can be said that $C$ is efficiently PAC learnable by the hypothesis class, $R$ over the distribution $D$. \newline \newline
\textit{Proposition 3.2.}\ If there exists a concept class $C$, such that it is \textit{R-evolvable} over distribution $D$, then it is also efficiently learnable using the statistical query model using $R$ over $D$.\newline \newline
\textit{Proof.}\ In the statistical query learning model, the PAC oracle, $EX(c,D)$, which gives random examples of the target concept,$f$ with respect to an input distribution $D$ over the domain, $X$ is replaced by a weaker oracle, say $SQ(c,D)$. The new oracle, $SQ(c,D)$ does not provide the learning algorithm, with individual random examples, but provides accurate probability estimates within arbitrary inverse polynomial additive error, over the sample space generated by the PAC oracle $EX(c,D)$.Say, we have a query $q=(q_{1},\alpha)$, where $q_{1}=q_{1}(x,l)$ is any Boolean function over inputs $x\in X$, such that $x$ is drawn from $D$ and $l\in \{0,1\}$, the oracle $SQ(c,D)$ will return a probability estimate that $q_{1}(x,c(x))=1$, which is accurate within additive error $\alpha \in [0,1]$.Hence, it can be said that the statistical query learning model is a weaker form of learning than Valiant's PAC model.\ The diagram below shows a block representation of statistical query learning, such that it acts as a messenger between the PAC oracle and the learning algorithm, which have no direct contacts between them.\newline
\begin{figure}[ht!]
\centering
\includegraphics[width=110mm]{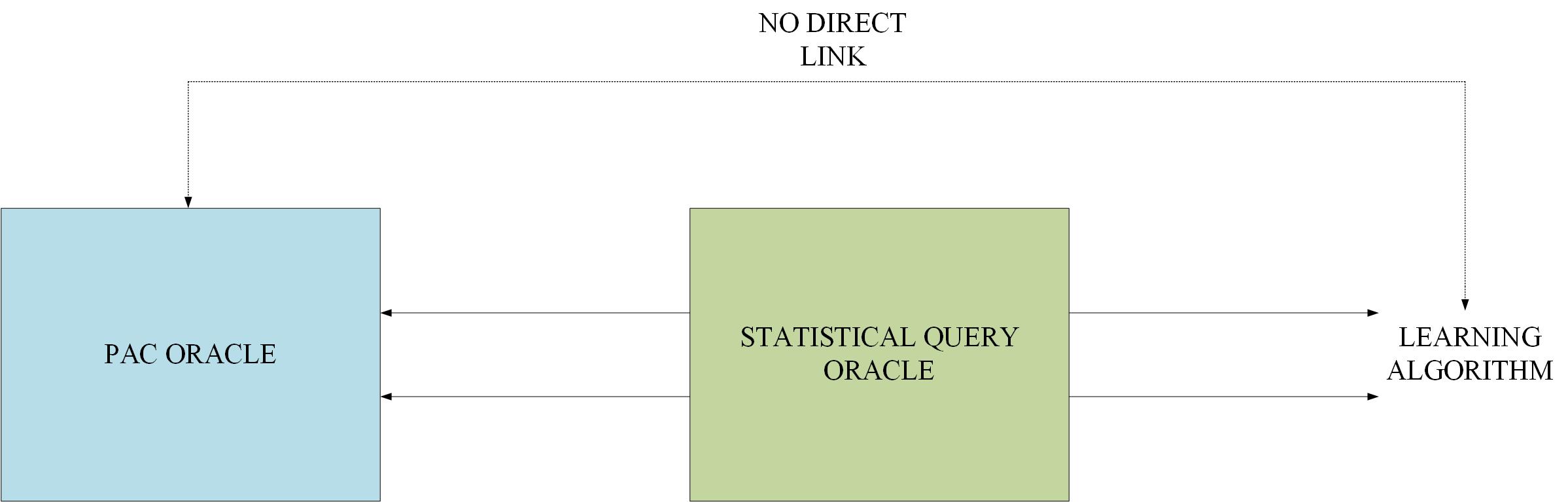}
\end{figure}
\newline
Let $r_{c}$ be a variant of the current representation $r\in R$, such that $r_{c}\in Neigh_{N}(r,\epsilon)$ and we want to measure the chances of $r_{c}$ being the next hypothesis of $r$, which is denoted by $Pr[r,r_{c}]$. So, we need pairs $(r,r_{c})$ from the set $POS$. In other words, we need to find the probability that for certain pairs $(r,r_{c})$, the empirical performance for the samples, whose size is upper bounded by $s$, follows $P(r_{c})\geq P(r)+t$. Let there be four events,whose corresponding queries are denoted by $q_{i}$ and additive errors by $\alpha_{i}$, for all $i\in [4]$. If the ideal function is $f$, the four events are $(r=f,r_{c}=f),(r=f,r_{c}\not=f),(r\not=f,r_{c}=f)$ and $(r\not=f,r_{c}\not=f)$ respectively for $i=1,2,3,4$. Each query, $q_{i}$ is a request for the probability of event $i$ on the distribution generated by the PAC oracle, $EX(f,D)$. Hence, a query $(q_{i},\alpha_{i})$ is interpreted as a request for the value $P_{x}^{q_{i}}=P_{i}(say)=Pr_{x\in D}[q_{i}(x,f(x))=1]=Pr_{EX(f,D)}[q_{i}=1]$ and since, $\alpha_{i}$ is the additive error of the probability estimates, so the statistical query oracle, $SQ(f,D)$ actually returns $\hat{P_{i}}$, where we have $P_{i}-\alpha_{i}\leq \hat{P_{i}}\leq P_{i}+\alpha_{i}$. Since, each $\alpha_{i}$ is bounded by inverse polynomial, so we have $1/\alpha_{i}=O(p_{\alpha}(1/\epsilon,n,size(f)))$, for some polynomial $p_{\alpha}$. Now, let us write the equations below, for the four events to give a clear picture.
\begin{equation*}
P_{1}=Pr_{EX(f,D)}[q_{1}=1]=Pr_{EX(f,D)}[r(x)=f(x)\ and\ r_{c}(x)=f(x)]
\end{equation*}
\begin{equation*}
P_{2}=Pr_{EX(f,D)}[q_{2}=1]=Pr_{EX(f,D)}[r(x)=f(x)\ and\ r_{c}(x)\not=f(x)]
\end{equation*}
\begin{equation*}
P_{3}=Pr_{EX(f,D)}[q_{3}=1]=Pr_{EX(f,D)}[r(x)\not=f(x)\ and\  r_{c}(x)=f(x)]
\end{equation*}
\begin{equation*}
P_{4}=Pr_{EX(f,D)}[q_{4}=1]=Pr_{EX(f,D)}[r(x)\not=f(x)\ and\ r_{c}(x)\not=f(x)]
\end{equation*}

Again for each additive error, $\alpha_{i}$, such that $1/\alpha_{i}=O(p_{\alpha}(1/\epsilon,n,size(f)))$, we have the following equations below.
\begin{equation*}
P_{1}-\alpha_{1}\leq \hat{P_{1}}\leq P_{1}+\alpha_{1}
\end{equation*}
\begin{equation*}
P_{2}-\alpha_{2}\leq \hat{P_{2}}\leq P_{2}+\alpha_{2}
\end{equation*}
\begin{equation*}
P_{3}-\alpha_{3}\leq \hat{P_{3}}\leq P_{3}+\alpha_{3}
\end{equation*}
\begin{equation*}
P_{4}-\alpha_{4}\leq \hat{P_{4}}\leq P_{4}+\alpha_{4}
\end{equation*}

Hence, we can say that the statistical query oracle generates probabilities as its output over the random sample space of the PAC oracle. It is very easy to simulate the oracle, $SQ(f,D)$ on some query, say $(q_{i},\alpha_{i})$, with probability at least $1-\epsilon$. A sufficient number of random labeled examples\ $\langle x,f(x)\rangle$, polynomial in $n$ and $1/\epsilon$ can be drawn from $EX(f,D)$, such that we only use the fraction of examples for which $q_{i}=1$ as the estimate, $\hat{P_{i}}$ of $P_{i}$.(Note that we always assume that the number of calls to the PAC oracle is bounded by the polynomial $p_{\alpha}$ and each $q_{i}$ is evaluatable in polynomial time, so that the efficiency is maintained). Thus, if the learning algorithm is given access to $SQ(f,D)$, it can be easily simulated given access to the PAC oracle, $EX(f,D)$.So, we can say that, the statistical query oracle is some sort of a midman between the learning algorithm and the PAC oracle, such that the learning algorithm has access to the statistical query oracle, which in turn uses the PAC oracle but there does not exist any direct contact between the algorithm and the PAC oracle.\newline \newline
Thus, the proof of the proposition is complete. \newline \newline
Follow the diagram below to have an intuitive understanding of the mechanism discussed above\ (all the $\hat{P_{i}}^{,}$s are replaced by $P_{i}$ for simplicity).
\begin{figure}[ht!]
\centering
\includegraphics[width=100mm]{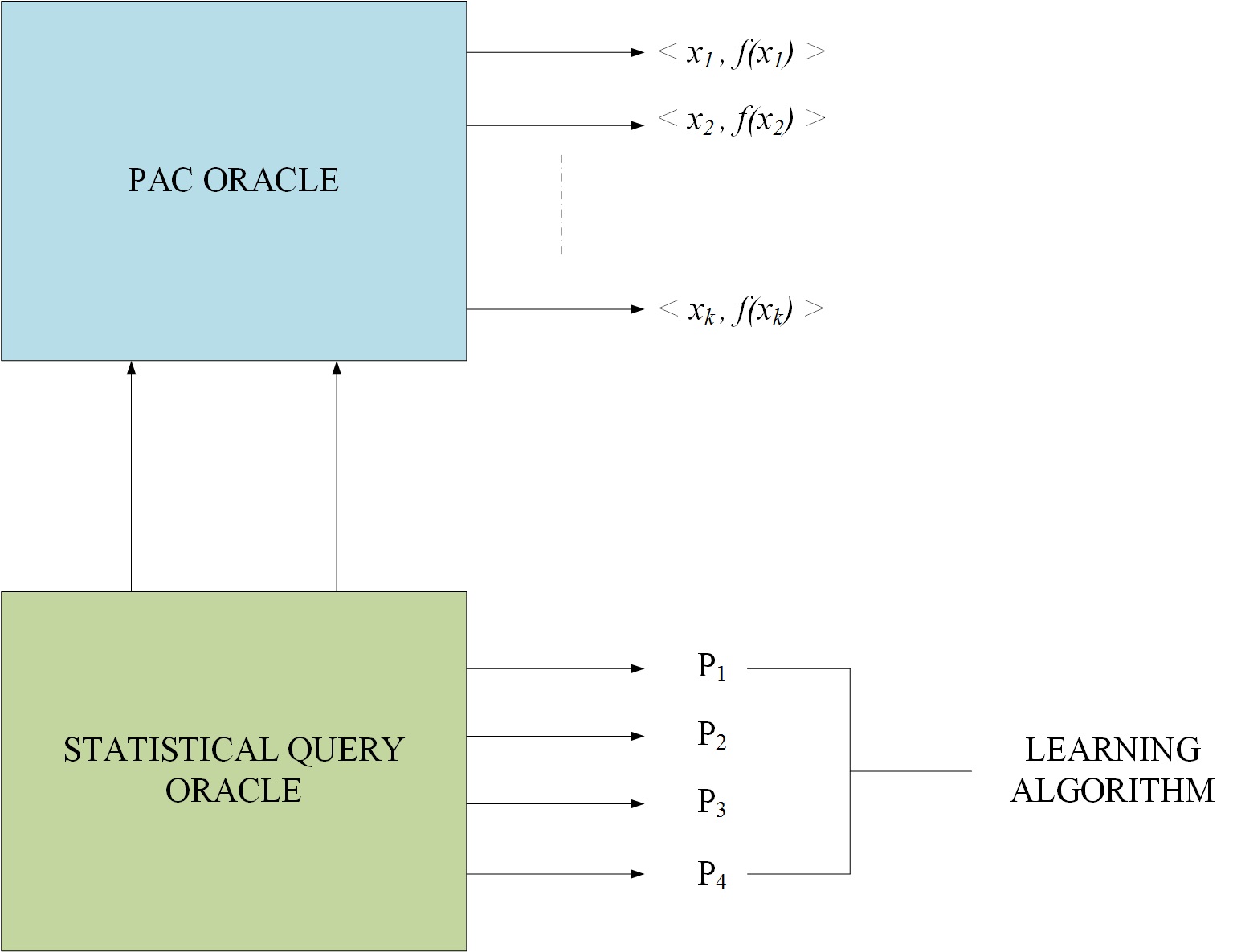}
\end{figure}
\newline
So, uptil now, we discussed whether a class is learnable or not, provided it is evolvable. But,what happens if we say the converse, i.e whether a class is evolvable or not, provided that it is learnable? Well, it is for sure, by simple intuition that if a class is not learnable, it can't be evolvable, which we will see in the following two propositions, with respect to the classes of parity functions and Boolean threshold functions.\newline \newline
\textit{Proposition 3.3.}\ If $F_{n}$ is the class of all parity concepts over $n$ Boolean variables, then there does not exist any polynomial time efficient statistical query learning algorithm for the class $F$, where $F=\bigcup_{n\geq 1}F_{n}$. \newline \newline
\textit{Proof}.\ The proof has been analyzed from [3].
\ Consider a query, $q:\{0,1\}^{n}\times \{0,1\}\rightarrow \{0,1\}$, which takes a variable $x\in \{0,1\}^{n}$ and $f(x)\in \{0,1\}$ as it input, to give a Boolean output. (note that $f$ here denotes the parity concept and not the ideal function in the model of evolution)\ Let us have our target distribuion, $D$, uniform over $\{0,1\}^{n}$ and represent the probability of the query that it is equal to 1, on an input generated by the PAC oracle $EX(f,D)$, such that $f$ is drawn randomly from $F_{n}$ by the variable, $P_{q}$. So, we have $P_{q}=Pr_{EX(f,D)}[q=1]$. As we are considering uniform distribution over $\{0,1\}^{n}$, it is easy to see that we have $P_{q}=(1/2^{n})\sum_{x\in \{0,1\}^{n}}q(x,f(x))$. Hence, we have the following equation below if we take expectations on both sides of the previous equation. 
\begin{equation*}
E[P_{q}]=\frac{1}{2^{n}} E[\sum_{x\in \{0,1\}^{n}}q(x,f(x))]
\end{equation*}
Using the additive property of expectations\ $(E[\sum k]=\sum E[k])$, we have the following equation below.
\begin{equation*}
E[P_{q}]=\frac{1}{2^{n}}\sum_{x\in \{0,1\}^{n}}E[q(x,f(x))]
\end{equation*}
Let the cardinality of the set, say $S_{0}$ consisting of all $x$, such that $q$ is always zero (independent of $f^{\ '}$s value) be equal to $c_{0}$ and that of the set, say $S_{1}$ consisting of all $x$, such that $q$ is always 1 ((independent of $f^{\ '}$s value))be equal to $c_{1}$.This means, when we have some $x\in S_{0}$, then $E[q(x,f(x))=0$ and when we have some $x\in S_{1}$, then $E[q(x,f(x))]=1$.\ Let $c_{2}$ be the the cardinality of the set, say $S_{2}$, where $S_{2}$ contains all such $x$, for which we have $q(x,0)\not=q(x,1)$.\ Thus, we get 
\begin{equation*}
E[P_{q}]=\frac{1}{2^{n}}\left(c_{1}+\frac{1}{2}c_{2}\right)
\end{equation*}
Again, we have
\begin{equation*}
E[P_{q}^{2}]=\frac{1}{2^{2n}}\sum_{x,y\in \{0,1\}^{n}}E[q(x,f(x))q(y,f(y))]
\end{equation*}
Intuitively it can be written as,
\begin{equation*}
E[P_{q}^{2}]=\frac{1}{2^{2n}}\sum_{x\in S_{0},y\in \{0,1\}^{n}}E[q(x,f(x))q(y,f(y))] +\ \frac{1}{2^{2n}}\sum_{x\in S_{1},y\in \{0,1\}^{n}}E[q(x,f(x))q(y,f(y))]\ + 
\end{equation*}
\begin{equation*}
\frac{1}{2^{2n}}\sum_{x\in S_{2},y\in \{0,1\}^{n}}E[q(x,f(x))q(y,f(y))]
\end{equation*}
After analyzing the results for each set, we finally get
\begin{equation*}
E[P_{q}^{2}]=\frac{1}{2^{2n}}(c_{1}(c_{1}+1/2\ c_{2})+c_{2}(1/2+1/2\ c_{1}+1/4\ (c_{2}-1)))
\end{equation*}
Again, squaring both the sides of the equation for $E[P_{q}]$, we get
\begin{equation*}
E[P_{q}]^{2}=\frac{1}{2^{2n}}\left(c_{1}+\frac{1}{2}c_{2}\right)^{2}
\end{equation*}
Hence, the variance can be calculated as,
\begin{equation*}
Var[P_{q}]=E[P_{q}^{2}]-E[P_{q}]^{2}=\frac{c_{2}}{2^{2n+2}}\leq \frac{1}{2^{n+2}}
\end{equation*}
Hence, for any query, $q$, the variance of the quantity, $P_{q}$ is exponentially small with respect to the random draw of the target concept and thus, it can be proved by contradiction, by taking the parameter $\epsilon$ to be any constant lesser than 1/4, fixing the parameter for $\alpha$ and then showing that a randomly chosen parity concept will be consistent with the query responses received by the learning algorithm, say $A$, by using Chebyshev's inequality and the bound for variance, calculated above.\ Hence, the error of $A$'s hypothesis must be large with respect to the random draw of the target concept, as many parity concepts are consistent with the responses received by $A$ and this, proves that the class of parity concepts is not efficiently SQ learnable.
\section{Evolvability of the class of monotone conjunctions}
In this section, we will show that the class of monotone conjunctions is evolvable, under some conditions, unlike that of the parity class and the class of Boolean threshold functions.\newline \newline
\textit{Theorem}.\ \ The class of monotone conjunctions is provably evolvable over the uniform distribution for their natural representations.\newline \newline
\textit{Proof}.\ \ Let us have a total of $n$ Boolean variables, say $x_{1},x_{2},...,x_{n}$. A monotone $k-$ conjunction is a conjunction of at most $k$ non-negated literals over the $n$ Boolean variables.\ If we want to apply our discussed evolution model to this case, it is very important that we have proper notion of neighborhood $N$, value of tolerance, $t$ and the sample sizes, before we evaluate the mutator random variable at each step.\ We take our hypothesis class (i.e the representation class) $R$ to be the class of all monotone $k-$ conjunctions, such that we want an accuracy of $\epsilon$ on the evolution of this class.For some positive constant $c$, fix the value of $k$ to be equal to $\lfloor \log (cn/\epsilon)\rfloor$. (We will see later, why we assumed this value)\newline
Let for any representation $r\in R$, we have a set $S^{r}$, which contains all the sets of conjunctions of literals in $r$. Then we have to either remove some literal from $r$ or add something to it or maybe add and then remove something to get a better hypothesis (representation), which has performance better than $r$ and gets one step closer to realizing the ideal function, $f$.\ So, it is logical to assume that the neighborhood $N$, which is represented by the notation, $Neigh_{n}(r,\epsilon)$, encompasses all these cases and thus, we get some better variant of $r$ from this set.\ Let us have the following sets below.\newline \newline
$S^{r}_{1}$\ -\ Sets of all the conjunctions which consist all the literals of $r$, with one extra literal added \newline
$S^{r}_{2}$\ -\ Sets of all the conjunctions which consist all the literals of $r$, with one literal removed \newline
$S^{r}_{3}$\ -\ Sets of all the conjunctions which consist all the literals of $r$, with one extra literal added and then another literal removed\newline \newline
So, we have $Neigh_{N}(r,\epsilon)=S^{r}_{1}\cup S^{r}_{2}\cup S^{r}_{3}$. Thus, either we can stay at $r$ and add or remove a literal to $r$ and then output a variant from them (this has a probability of 1/2)\ or we can add a literal to $r$ and then remove another literal from it, to get a better variant\ (this also has a probability of 1/2).\ Hence, from above, clearly $N$ is an $O(n^{2})$ neighborhood of $r$. Now, we have to discuss about the tolerance, sample size and performance issues of this construction.\newline \newline
(Note that we will represent the tolerance, $t(1/n,\epsilon)$ and the number of samples, $s(n,1/\epsilon)$ by simply $t$ and $s$ respectively.) \newline
Let us take $t=\frac{1}{2}.(2^{-2k})$ and $s=1/t^{3}$.\ Since, we have taken $k=\lfloor \log (cn/\epsilon)\rfloor$, so we get $t=\frac{\epsilon^{2}}{2c^{2}n^{2}}$. The construction is done in such a way that every mutation in the neighborhood $N$ will cause a performance improvement of at least $2t$ or causes no improvement. The test, which has been devised identifies the right mutation except with some small exponential probability (which we get by Hoeffding bound) from the one which gives no improvement in performance.Recall the theorem of the Hoeffding bound. It says that the probability that the mean of $s$ independent random variables , with each taking values in the range $[a,b]$, is greater than or less than the mean of their expectations by more than $\delta$ is at most equal to $e^{\frac{-2s\delta^{2}}{(b-a)^{2}}}$. In our case, we have $a=-1$ and $b=1$, which we have seen before. Again, since the variable $\delta$ captures deviations, so it must be equal to the tolerance, $t$. So, we have $\delta=t$ and hence, the number of trials, which is nothing but the sample size, $s$ is equal to $1/\delta^{3}$. So, by Hoeffding bound, the probability that all the $s$ mutation trials each with an expected improvement of $2\delta$ will produce a mean improvement of less than $t=\delta$ is at most equal to $e^{\frac{-2.\frac{1}{\delta^{3}}.\delta^{2}}{4}}=e^{\frac{-1}{2\delta}}=e^{\frac{-1}{2t}}=e^{-\frac{c^{2}n^{2}}{\epsilon^{2}}}$, which is indeed very small.\ Now, if we take $g(n,1/\epsilon)$ to be the number of stages (equivalent to saying number of generations) and in each stage, we have a total of $p$ variants, which are to be tested, so if we calculate $g.p.e^{-\frac{c^{2}n^{2}}{\epsilon^{2}}}$,we see that it is strictly less than $\epsilon/2$ if we fix some $c>\frac{\epsilon}{n}\sqrt {\frac{\ln 2pg}{\epsilon}}$ properly, which is indeed what we want.\newline \newline
Now, finally we prove a set of claims about the testing and performance by 
modifying and manipulating the current representation, $r$ as a whole and try to give combined short proofs of them, according to [1].\ We say, that $r$ is a monotone conjunction of $m$ literals, where $m\leq k$. Consider $r=y_{1}y_{2}...y_{m}$, where each $y_{i}\in \{x_{1},x_{2},...,x_{n}\}$.\ Let $A$ be the set of $v$ literals, which form the conjunction of the ideal representation, $f$ and $B$ be the set of $m$ literals, which form the conjunction of the current hypothesis $r$.\ So, let us have a set of cases and try to prove them. \newline \newline
\textit{Case\ 1}.\ If we have $m<k$, i.e the number of literals in the current hypothesis is strictly less than that of the ideal conjunction, then intuitively we can see that if we add a literal from the true function to the hypothesis, then our performance will increase and it is indeed so, mathematically.\ So, consider a literal, $l$ in the set $A-B$, which will be added to the present hypothesis, $r=y_{1}y_{2}...y_{m}$. Hence, adding $l$ to $r$ will change the value of the hypothesis from +1 to -1 on the points which satisfy the conjunction, $l^{'}y_{1}y_{2}...y_{m}$ and thus, $l^{'}=1$, which makes $l=-1$ and so, the ideal function must be equal to -1 on these points as we have $l=-1$ and $l\in A$.\ As, we know that over the uniform distribution, $U$ the probability that a conjunction of $d$ literals will be satisfied is equal to $2^{-d}$\ (and hence for a disjunction to be satisfied, it is equal to $1-2^{-d}$), so in our case, the $(m+1)$ points in $r$ will have a probability of $2^{-(m+1)}\geq 2^{-k}$. Thus, performance increase will be at least equal to $2.2^{-k}=2^{1-k}$. (notice here that the value of the hypothesis is equal to the target concept, i.e $r=f$ and hence, the performance is bound to increase).\newline \newline
\textit{Case\ 2}.\ Consider a literal, $l$ in $A\cap B$, which will be removed from $r$.\ Now, since the literal is also present in the ideal function, so it will surely decrease the performance.\ Let us take $l$ to be equal to $y_{1}$, without loss of generality.Hence, the hypothesis changes the value from -1 to +1, such that it satisfies the conjunction $l^{'}y_{2}y_{3}...y_{m}$ and thus, $l=-1$, which makes value of the ideal function to be equal to -1. Hence, the probability of the points in the hypothesis is equal to $2^{-m}\geq 2^{-k}$ and, since $r\not=f$, the performance value decreases by at least $2.2^{-k}=2^{1-k}$. \newline \newline
\textit{Case\ 3}.\ Consider two literals, $l_{1}\in A-B$ and $l_{2}\in B-A$, such that we add $l_{1}$ to $r$ and then remove $l_{2}$ from $r$.\ Now, from the previous two cases, we see that adding $l_{1}$ to $r$ will change the hypothesis from incorrect to correct and has a probability of $2^{-(m+1)}$ and removing of $l_{2}$, also applies with a probability of $2^{-(m+1)}$ and the change in some of the points is from correct to incorrect. Now, we have to show that the net change is not neutral, but positive for the evolution performance, which is very easy.\ Consider the set of literals, $L$, such that they are missing from $r$, but are present in $f$ (of course, other than $l_{1}$) and we have $|L|=x$. Now, if we consider that $l_{2}$ was removed, so we have a satisfying conjunction of $l_{1}{l_{2}}^{'}y_{2}...y_{m}$ and it is also known that $x$ fraction of these have a value of 1 in the ideal function, with a probability equal to $2^{-x}$.\ Thus, the improvement in performance is at least equal to $2^{-x-m}\geq 2^{-|A|-k}$.\newline \newline
\textit{Case\ 4}.\ Let us have two literals, $l_{1}\not\in A$ and $l_{2}\in A\cap B$, so that $l_{1}$ is added to $r$ and $l_{2}$ is removed from $r$.\ If we look carefully into this, we can see that it is exactly the opposite of the previous case.\ Now, removing $l_{2}$ is an incorrect change at every point, as it belongs to both the sets $A$ and $B$ and occurs with probability, $2^{-(m+1)}$.\ Consider the set of literals, $L$ such that they are absent from the ideal function, $f$ and let $|L|=x$.\ Then we can find that on the domain of points satisfying, ${l_{1}}^{'}y_{1}y_{2}...y_{m}$, $x$ fraction of them have a correct value of 1 on the ideal function, with a probability of $2^{-x}$, which decreases the performance measure, by at least $2^{-x-m}\geq 2^{-|A|-k}$.\newline \newline
\textit{Case\ 5}.\ Analyzing the above four cases, we find that the performance remains constant when we combine the addition of a literal in $A-B$ to $r$ from Case 3 with the removal of a literal in $A\cap B$ from $r$ and the same thing happens when we combine the addition part of Case 4 with the removal part of Case 3.\ Consider two literals $l_{1}\in A-B$ and $l_{2}\in A\cap B$. From these two, it is clear that $l_{1}$ belongs to the ideal function, but does not belong to the current hypothesis. Again, $l_{2}$ belongs both to the target function and the current hypothesis. As we analyzed in the previous cases, the addition of $l_{1}$ to $r$ is a correct change at every point and the probability is equal to $2^{-(m+1)}$. Also, the removal of $l_{2}$ from $r$ is an incorrect change at every point and has an equal probability of $2^{-(m+1)}$.\ So, the effect of two changes gets cancelled.\ The proof of the second case is similar.\newline \newline
\textit{Case\ 6}.\ If the current representation, $r$ contains all the literals that are already present in the ideal conjunction, $f$ and it is logical, that if we remove all the irrelevant variables in $r$, which are not present in the target, it will increase the performance measure.(\textit{It seems that this should be the final step of evolution, in which it has been able to learn the target genome and now is ready to get rid of all the unworthy life experiences and mutate to an advanced life form. It is indeed the case, which we will see later})\newline
Without loss of generality, assume $l_{1}=y_{1}$.\ Now, removing $l_{1}$ from $r$, will change the hypothesis's value from -1 to +1 on the points satisfying ${l_{1}}^{'}y_{2}y_{3}...y_{m}$. Again, since $r$ contains all the literals in the target and all such points have a true value of +1 with a probability of at least $2^{-m}$, hence the performance must increase by at least $2.2^{-m}=2^{1-m}\geq 2^{1-k}$.\newline \newline
\textit{Case\ 7}.\ Have a literal, $l_{1}$ in the set, $B-A$, which is to be added to the current representation, $r$, which already contains all the literals of $f$.\ Therefore, comparing with the previous cases, the value of the hypothesis changes from +1 to -1 on the points satisfying ${l_{1}}^{'}y_{1}y_{2}...y_{m}$. Similar to previous case, if $r$ contains all the literals in $f$, then all such points must have true values of +1.\ Hence, we got a disagreement between the hypothesis and the target concept, and so the performance must decrease by $2^{-m}\geq 2^{-k}$.\newline
(\textit{If we consider the intuitive thinking of this case, it is like, that the evolution mechanism is already prepared to mutate and getting rid of the unwanted variables, but some extraneous agent is trying to stuff an unwanted life experience to prevent the process of mutation, by decreasing its performance measure}).\newline \newline
\textit{Case\ 8}.\ What happens when we have $|A|>k$, i.e the number of literals in the ideal function is greater than the number of literals in the initial representaton, of the whole evolution process?\ Well ,then the fraction of points on which the prediction of -1 is there, increases by $(1-2^{-m-1})-(1-2^{-m})=2^{-m-1}$.\ Now, if we have $m\leq k-2$, then the fraction increases by at least $2^{1-k}$ and a literal is added.\ The fraction may also decrease by at least $2^{1-k}$, with the removal of one literal if we have $m\leq k-1$.\ Hence, for the case when we have $|A|>k$, the corresponding increase or decrease in the fraction of points on which the prediction is correct is at least $2^{-k}$. \newline \newline
Thus, according to the above mentioned rules, the evolution mechanism functions and tries to have the best possible representation, closest to the ideal function.\ So, it only chooses the cases, where there seems to be an increase of performance by adding or deleting literals to or from the conjunction of the current representation.\ Now, it can be possible that some non-ideal literals may have been added to the representaion, without decreasing the performance, but if the ideal conjunction has a total of say, $z$ literals, then by the coupon collector's problem, after a total of $(n\log z+n\log 1/\epsilon)$ generations, all $z$ would have been swapped or added in, thus giving the final genome, except with a probability of $\epsilon$.\ Thus, [1] says that if the initial number of literals in the representation, $r_{0}$ and the number of literals in the ideal conjunction is at most equal to $k$, then the evolution mechanism gets completed in the above mentioned number of stages (or generations), except with a probability of $\epsilon$, where $\epsilon$ is the error parameter due to various factors, hindering evolution.\ Again, if we have $m>k$, then the removal of any literal from the hypothesis will change the value on at most, $2^{-m}<2^{-k}=2\epsilon/cn$ of the distribution and hence, the performance can decrease at the most by $2\epsilon/cn$.\ Thus, if we set the tolerance, $t$ to $4\epsilon/cn$ and hence, we have $\delta=2\epsilon/cn$ for the Hoeffding Bound analysis, for checking the probability when we decrease the number of literals from the hypothesis, we actually have a neutral condition of mutation. But, putting $a=-1,b=1,\delta=2\epsilon/cn=t/2$ and $s=\delta^{3}$, the probability is at most $e^{-cn/4\epsilon}$, which is very less.\ Thus, after running the process for $O(n)$ stages, we will be able to reach to a smaller conjunction of length $k$, except with an exponentially small probability.\ We can also show optimality for the 8th case, by similar manner.\newline \newline
By duality principle, we can convert a conjunction to a disjunction, by the laws of negation and applying De-Morgan's laws, we can also prove evolvability for the class of disjunctions.

\section{Critical assessment, suggestions and open problems}
I discuss some open problems and further suggestions to extend the model of evolution of [1]. Though [1] is undoubtedly one of the very few best papers in this area of evolution and theoretical computer science, but according to me, there is still much to be done in this new field of study.\ Let us discuss some of them below.\newline \newline
\textbf{1}.\ We have some forceful assumptions in the definitions of evolvability, such that it falls in the framework of PAC learning model. Though, we can say that a genome is evolvable only when its performance is close to 1, i.e it, almost is same to the target function (target genome) and so can assume that the final performance is equal to $1-\epsilon$, for some small $\epsilon>0$, but is it necessary to bound the resources and time of computing the variants by polynomials. How can we be so sure, that everything is polynomially bounded?\ We are basically assuming beforehand that evolutionary mechanism is a PAC learning model in the definitions, that we define. Maybe, we could have assumed that the final performance is some $0<x<1$ and then established the PAC learnability of evolution. \newline \newline
\textbf{2}.\ Maybe, there exists an algorithm with nature which efficiently searches all the exponentially possible many variants of the current genome. Maybe, there are $c.2^{n}$ many representations, for some positive constant $c$, but somehow the evolutionary mechanism knows beforehand which representation to pick and then may apply binary search, so that the search completes in $O(n)$ time, which is very efficient.\ We should always understand that evolution mechanism is itself learning for millions of years and through its experience, it may have some information of which representation to select (or there may exist some probability distribution over the variants and there may exist many variants which are capable of evolution in the probability space) for the next generation and hence, can perform a binary search over all the variants. The idea of discarding most of the variants seems to be too much of an assumption.\newline \newline
\textbf{3}.\ It may be possible that evolution has some efficient algorithm to search all the exponential variants in efficient polynomial time and finding the algorithm, will be one of the the first stepping stones towards the great P-NP problem from biological evolution point of view.\ It may be probable that nature has some algorithm, which outputs the evolved genome without even looking at all the variants. (Here where, statistical query learning may have great importance to formulate a better theory of evolution, by generating a favorable probability space over the variants). It is the right time, that computer science theorists and mathematicians look more deeply into Darwinian theories, to some how come close to one of the greatest problems ever posed by the humans.\newline \newline
\textbf{4}.\ Does there exist any distribution free evolution model, is a very general and broad question and it may take quite a long time to get a satisfactory answer to this question.[1] \newline \newline
\textbf{5}.\ Finally, it is unknown whether we can PAC learn the class of DNF using DNF hypothesis and find a consistent hypothesis finder in polynomial time.\ So, there is some possibility that we learn the evolvability of the DNF expressions, where we start our initial representation of the genome with a DNF and the target concept is also a DNF.\ Since, DNFs are extremely rich in their representations and can entail lots of valuable information, so it is possible that evolution mechanism uses this class of expressions and still evolves into a stronger individual within a finite number of generations.\ Hence, evolvability with respect to the class of DNFs, will be an open problem for a great amount of time, according to me.

\bibliographystyle{splncs}

\end{document}